\documentclass[twocolumn]{article}
\usepackage{cite}
\usepackage{amsmath,amssymb,amsfonts} 
\usepackage{graphicx}
\usepackage{textcomp}
\usepackage{diagbox}
\usepackage{multirow}
\def\BibTeX{{\rm B\kern-.05em{\sc i\kern-.025em b}\kern-.08em
    T\kern-.1667em\lower.7ex\hbox{E}\kern-.125emX}}

\usepackage[top=1.5cm,bottom=2cm,left=1.3cm,right=1.3cm]{geometry}

\usepackage[switch]{lineno}

\usepackage{hyperref}
\usepackage{graphicx} 
\usepackage{amsmath}
\usepackage{amssymb} 
\usepackage{rotating} 
\usepackage[mathscr]{euscript}
\usepackage[normalem]{ulem}
\usepackage{authblk}

\DeclareMathOperator*{\argmin}{argmin}

\usepackage{subcaption}

\usepackage[utf8]{inputenc}

\begin{document}

\title{Simultaneous Multi-View Camera Pose Estimation and Object Tracking with Square Planar Markers} 
\date{}
\author[1]{Hamid Sarmadi}
\author[2,1]{Rafael~Mu\~noz-Salinas}
\author[3]{M.A. Berb\'is}
\author[2,1]{R. Medina-Carnicer}

\affil[1]{Instituto Maim\'onides de Investigación Biomédica de C\'ordoba (IMIBIC), C\'ordoba,  Spain}
\affil[2]{Departamento de Informática y Analisis Numérico, Edificio Einstein, Campus de Rabanales, Universidad de Córdoba, C\'ordoba, Spain}
\affil[3]{Grupo Health Time, Avda Brillante 106, Córdoba, Spain}

\maketitle

\begin{abstract}
Object tracking is a key aspect in many applications such as augmented reality in medicine (e.g. tracking a surgical instrument) or robotics. Squared planar markers have become popular tools for tracking since their pose can be estimated from their four corners. While using a single marker and a single camera limits the working area considerably, using multiple markers attached to an object requires estimating their relative position, which is not trivial, for high accuracy tracking. Likewise, using multiple cameras requires estimating their extrinsic parameters, also a tedious process that must be repeated whenever a camera is moved.

This work proposes a novel method to simultaneously solve the above-mentioned problems. From a video sequence showing a rigid set of planar markers recorded from multiple cameras, the proposed method is able to automatically obtain the three-dimensional configuration of the markers, the extrinsic parameters of the cameras, and the relative pose between the markers and the cameras at each frame.  Our experiments show that our approach can obtain highly accurate results for estimating these parameters using low resolution cameras.

Once the parameters are obtained, tracking of the object can be done in real time with a low computational cost. The proposed method is a step forward in the development of cost-effective solutions for object tracking.

\end{abstract}


\maketitle
\section{Introduction} 
Square planar markers, comprised of a black external border and an inner binary pattern have gained popularity since their pose can be estimated by only using its four corners. As a consequence, they are being employed in 6DoF tracking tasks in different disciplines such as human-computer interaction \cite{wu_dodecapen:_2017}, augmented/virtual reality \cite{nee_ar_applications_2012}, robotics \cite{fiala_robot_2004} or in medicine for tracking of surgical instruments \cite{Wang:2015,NAKAWALA201850,bone_tumour,zhang_real-time_2017},  a crucial process of image-guided surgery providing the  position and orientation (pose) of the instruments with respect to the patient in preoperative registration and intraoperative navigation. 

Tracking one marker with only one camera imposes strong limitations on the effective working area. Therefore, many authors have employed several markers attached to the device to improve visibility, and/or multiple cameras to increase the tracked area. However, this involves two problems:  estimating the rigid structure of the marker set and the relative pose of the cameras.

Obtaining the structure of the marker set can be a time-consuming process prone to errors. Thus, many authors design their marker object using basic shapes like cubes or hexagons and obtain their relative positions from the equations of such shapes.  In order to estimate the relative poses of the camera set, most people employ a planar chessboard. However, when a circular camera configuration is employed (i.e., cameras placed in a circle pointing to its center), the chessboard is not simultaneously seen by all cameras. Then, it is required to acquire multiple views of the chessboard to find the relative camera poses. It is a time-consuming process that needs to be repeated whenever a camera is moved. 

This work proposes a method that given a synchronized video sequence showing an object comprised by a set of squared planar markers moving freely in front of the cameras, automatically estimates the 3D structure of the planar markers (object), the 3D poses of the cameras, and the relative pose between the object and the cameras.  Once the relative pose of the marker and the cameras have been obtained from the video sequence, the estimation of the pose between the two can be done in real-time by minimizing the global reprojection error of the observed marker corners in all the camera images. 

Our approach is a step forward in the use of planar markers for rigid object tracking. First, it allows tracking devices of any shape by simply attaching affordable markers, without paying attention to the position in which they are placed. Second, it allows to easily reconfigure the camera distribution to adapt to the requirements of the application.

This is, up to our knowledge, the first method in the literature that solves the three problems simultaneously and is publicly available \footnote{ \url{https://www.uco.es/investiga/grupos/ava/node/60}} for other researchers.

The remainder of this paper is structured as follows. Section~\ref{sec::relwork} explains the most related works. Section~\ref{sec::proposed_method} describes the proposed method while Section~\ref{sec::results} details the experiments conducted to validate our proposal. Finally Section~\ref{sec::conclusions} draws some conclusions.

 \begin{figure}[t]
    \centering
     \includegraphics[width=0.45\textwidth]{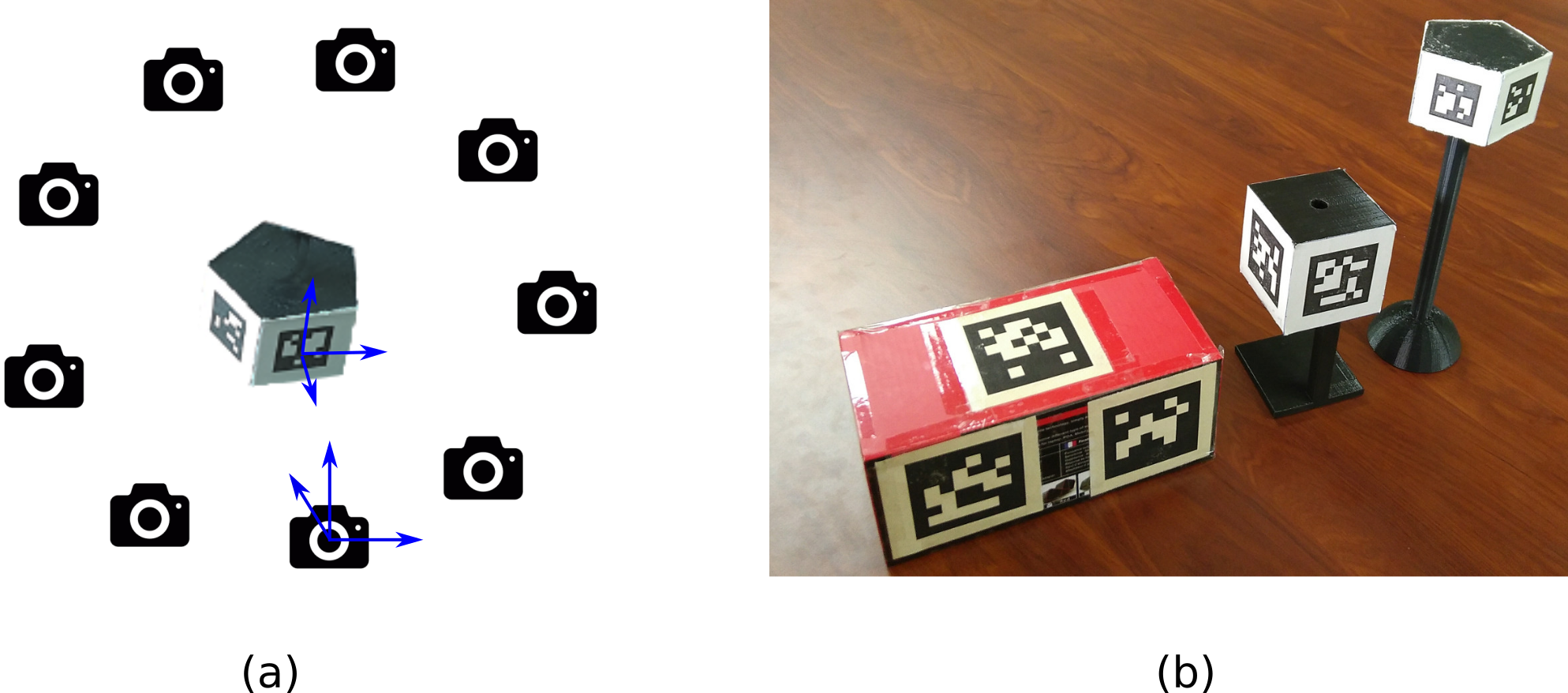}
    \caption{ Typical scenario for our method. (a) Cameras in a circular configuration tracking an object comprised by a set of markers. Blue axis represents the reference camera and marker. (b) Different types of object created by attaching planar markers. }
     \label{fig::teaser}
\end{figure}

\section{Related Work}
\label{sec::relwork}

\subsection{Squared planar marker tracking}

Object tracking requires finding correspondences between known points of the object and their camera projections. While some methods seek natural features such as key points or textures \cite{orb-slam,PTAM},  squared planar markers are an attractive solution because they are easy to detect and allow to achieve high speed and precision \cite{Aruco2014,artagPAMI,artoolkit,studierstube,GarridoJurado2015}. They are composed of an external black border and an internal code (most often binary) that uniquely identifies them. Also the corners of a single marker can be employed for camera pose estimation \cite{Aruco2014}.  

For example, Nakawala~{\em et al.} \cite{NAKAWALA201850} develop a surgical training system for Thoracentesis, where markers are attached to different instruments/materials such as syringe and catheters. Matthies ~{\em et al.} \cite{Matthies2015}  show how the use of squared planar markers can be employed to alleviate the line-of-sight problems of infrared tracking systems in Computed Tomography problems. For that purpose, they employ a cube of markers that are tracked with a camera. 
The authors of \cite{Kanithi:2016:IAR:3009977.3010023} create an Augmented Reality system that uses planar markers to track the trajectory of a needle in ultrasound-guided interventions.

Continuing in the surgical area, the work of Enayati {\em et al.}  \cite{Enayati2017} presents a framework for shared-control teleoperation of robotic arms such as the da Vinci robot. They employ a set of planar markers to delimit the controlling area on which the controller moves the robot by a haptic device.

Another interesting augmented reality application using planar markers is \cite{8088503}, which proposes a system to aid breast surgical planning. The system projects 3D ``holograms'' of images from breast MRI onto the patients using Microsoft HoloLens. The planar markers are employed to properly align the preoperative models with the real one.

Silvio Pflugi {\em et al.} \cite{10.1007/978-3-319-43775-0_8} propose a cost-effective navigation system for peri-acetabular Osteotomy Surgery using a Raspberry Pi that tracks a planar marker that can be directly attached to the patients' pelvis. They prove that their system shows no statistical difference compared to a much more expensive Polaris tracking camera.

In the context of bone tumor resectioning, the authors of  \cite{bone_tumour} evaluate the accuracy of augmented reality based navigation assistance in a pig femur model. As in previous cases, planar markers attached to the materials are employed to track their locations.

The work of Marcon {\em et al.} \cite{8265377} shows an approach to evaluate the posture of people by attaching several planar markers to their back. The authors focus on analyzing the activity of a dentist during a dental operation. As the authors indicate, the advantage of using markers is the absence of powered and/or heavy and cumbersome markers like wearable cameras or accelerometers.
 
Ghazi {\em et al.} \cite{ghazi_vision-based_2017} design bracelets made of planar markers sequentially attached to each other for the purpose of tracking hand and feet of infants. The bracelet was designed as a heptagon which allows a good visibility using even a single camera, but its dimensions could only be manually estimated. Wu {\em et al.} \cite{wu_dodecapen:_2017} use markers on a regular dodecahedron to track the movement of a pen and estimate its drawing. In these methods, however, there are strong assumptions about the relative poses of the markers with respect to each other.

Mu\~noz-Salinas {\em et al.} \cite{munoz-salinas_mapping_2018} map and track multiple markers fixed in an environment. They have a similar method to initialize marker poses on the map and make no assumptions about the relative poses of the markers with respect to each other. However, they use only one camera for mapping and localization.

Franciosa {\em et al.} \cite{franciosa_automatic_2011} employ a pair of stereo cameras to determine and track the 3D position of multiple planar markers. They do not, however, estimate the 3D pose of every marker but only the 3D location of the marker centers. 

\subsection{Camera calibration}
Extrinsic calibration of a set of cameras is a process already studied in the literature, and many solutions have been proposed. The problem consists in estimating the pose of a set of static cameras and the solution consists in obtaining references that can be matched between the different cameras.  Since cameras are in different locations, they only share a limited portion of the field of view. Then, the calibration process is normally done in clusters (normally camera pairs) and then all the information combined afterward. Some authors have employed 1D objects for extrinsic calibration, such as \cite{Svoboda2005} and \cite{Wang2007} for this task. The main problem is that in general this processes requires special equipment and/or settings such as active lighting. The most popular solution consists in using a  2D calibration pattern, such as \cite{BoLi2013}. Again, the main problem is that visibility of pattern is limited and it must be moved to different locations to be observed by all cameras. Indeed, whenever one of the cameras is moved, the process must be repeated.  The work Schmit et al. \cite{schmidt_calibration_2014} shows an interesting approach to the problem consisting in using a robot that moves the calibration pattern along the working area. In the recent work of Zhao {\em et al.} \cite{ZHAO201846}, the authors use both 2D chessboard calibration patterns and squared fiducial marker attached to the cameras in order to obtain the extrinsic parameters. Weng ~{\em et al.} \cite{Wang:2015} propose a non-iterative method to obtain the camera extrinsic parameter using a single fiducial squared marker.

In \cite{7504640} the authors employ squared fiducial markers in order to jointly calibrate an X-ray imaging system and several RGB cameras.
Approaches using  3D objects such as \cite{Wilson2017} and \cite{Shen2008} try to reduce the number of necessary images to do the calibration. 

The main problem with 3D objects is that obtaining its configuration is a complex process. The method proposed in this paper is very convenient since automatically obtains the marker configuration along with the camera extrinsic parameters. Unlike the previous approaches, ours can estimate the camera extrinsic parameters using a 3D object of unknown configuration.  This is, up to our knowledge,  the first method that simultaneously reconstructs the object configuration and the camera extrinsic parameters.

\section{Proposed Method}

 \label{sec::proposed_method}
We assume that there is a set of cameras that synchronously capture frames of a scene where a set of markers appear in different locations of the images.  Also, we assume that cameras do not move with respect to each other and the same is true for the markers. However,  the set of markers move w.r.t the camera set. Among the cameras, we pick one randomly as the reference camera and from markers we pick one as a reference marker; we use them to measure the relative 3D pose between the collection of the cameras and the collection of the markers (see Figure ~\ref{fig::teaser}a). 

Given an object or a scene, with the planar markers attached to it, we want to determine the relative poses between the cameras and the relative pose between the markers. We also want to find the relative pose between the set of cameras and the set of markers at each frame where at least one marker is detected.
 
 We start by detecting the planar markers in the 2D images captured by cameras throughout the video sequence. Then we try to estimate the 3D pose of the markers knowing the intrinsic parameters of the camera and the dimensions of the marker. When using real-world data, due to measurement errors, it is not always possible to confidently estimate the pose of a planar marker w.r.t to the camera. Due to small errors in the estimation of the corners, two possible poses arise as valid solutions (\cite{schweighofer_robust_2006}). This is the problem known as {\em pose ambiguity} in planar pose estimation. To address this issue, a confidence value is assigned to each solution so that in the final phase, the best one is considered.
 
\begin{figure*}[p]
    \centering
    \includegraphics[width=.8\textwidth]{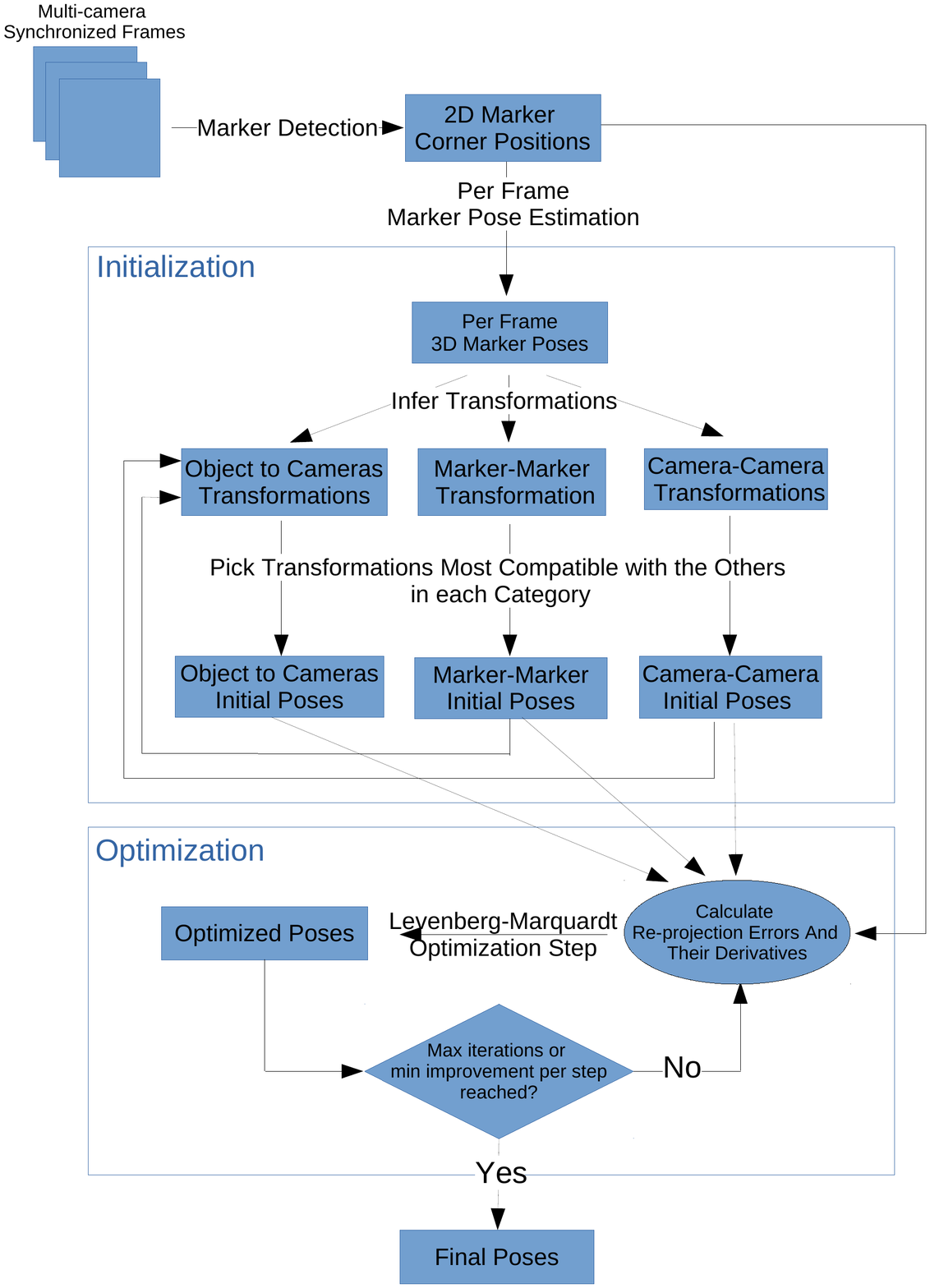}
    \caption{The flowchart of our algorithm.}
    \label{fig:flowchart}
\end{figure*}

We collect all the hypotheses for the relative position of the markers w.r.t to the cameras they are detected in. They are used to find an initial 3D pose for each camera w.r.t to the reference camera, an initial 3D pose for each marker w.r.t. the reference marker, and for each image, an initial relative pose between the reference camera and the reference marker. Finally, a non-linear optimization method is employed to globally minimize the reprojection error. You can see a flowchart of our algorithm in FIGURE \ref{fig:flowchart}.

The ArUco method \cite{Aruco2014} is employed to generate and detect our square-shaped markers. It is assumed that the side length of all markers is the same and it is known.
 
Below, we provide a detailed explanation of the proposed method and the preliminary concepts required to understand the solution introduced.
\begin{figure}[t]
\centering
  \includegraphics[width=0.40\textwidth]{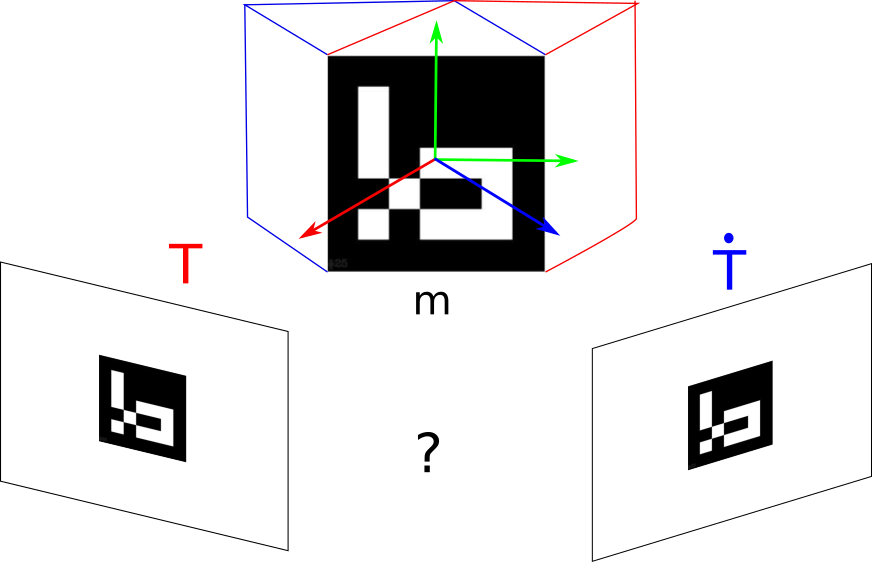}
\caption{Pose ambiguity problem: the same observed projection can be obtained from two different camera poses $T$ and $\dot{T}$. }
\label{fig:ambiguity1}
\end{figure}

\subsection{The ambiguity problem}
\label{sec::ambiguity_problem} 
The pose of a marker w.r.t. a camera can be estimated from its four corners. However, in practice, due to noise in the localization of the corners, two solutions appear, and in some cases, it is impossible to distinguish the correct one. The problem is depicted in Fig~\ref{fig:ambiguity1} where the marker, represented as one side of a cube, could be in two different orientations (red and blue color) thus obtaining almost identical projections from two different camera locations $\mathrm{T}$ and $\dot{\mathrm{T}}$. The methods proposed in \cite{Oberkampf1996495,Collins2014,schweighofer_robust_2006} find the best solution by a careful analysis of the projections. Most often, the reprojection error of one of the solutions is smaller than the reprojection error of the other one. In that case, there is no ambiguity problem and the correct solution is the one with the smallest error. However, when the marker is far from the camera, the errors in the estimation of the corners become relatively large. Then, the reprojection error of both solutions is so similar that it is not possible to determine the correct one. In this work, the ambiguity problem is considered so as to avoid errors in the optimization process.

 \subsection{Problem definition}
 Let us assume that there are $n$ cameras,  $q$ markers, and denote \begin{equation}
\mathcal{Q}=\{c\in\{1,\dots,n\}\},
\end{equation}
the sets of cameras, and
\begin{equation}
\mathcal{M}=\{m \in\{1,\dots,q\}\},
\end{equation}
the set of markers. In the set $\mathcal{Q}$ we choose a camera $c^*$ as the reference camera and similarly a reference marker $m^*$ from the set $\mathcal{M}$. 

Considering that markers are squares of side length  $s$,  let $\{u_m^{l}\}, l \in \{1,\ldots,4\}$ be the corners of a marker $m$ defined with respect to its own center as:
\begin{equation}
\begin{tabular}{l} 
$u^1_m$=( \begin{tabular}{rrr}~s/2,&-s/2,&0\end{tabular}),  \\
$u^2_m$=( \begin{tabular}{rrr}~s/2,&~s/2,&0\end{tabular}),\\
$u^3_m$=( \begin{tabular}{rrr}-s/2,&~s/2,&0\end{tabular}),\\
$u^4_m$=( \begin{tabular}{rrr}-s/2,&-s/2,&0\end{tabular}).\\ \end{tabular}
\end{equation}

 We define a transformation in the 3D space as a $4\times4$ matrix in the form of 
 $$\begin{pmatrix}
 \mathbf{R} & \mathrm t\\
 0 & 1
 \end{pmatrix},
 $$ 
 where $\mathbf{R}$ is a $3\times3$ rotation matrix and $\mathrm t$ is a $3\times1$ translation vector. To apply the transformation to a 3D point $p=(x,y,z)^\top$ we perform the following calculation:
 $$
 (x',y',z',1)^\top=
 \begin{pmatrix}
 \mathbf{R} & \mathrm t\\
 0 & 1
 \end{pmatrix}
(x,y,z,1)^\top
 ,$$
taking $p'=(x',y',z')^\top$ as the result.

Assuming that the cameras are fixed with respect to each other along the video sequence, let $\mathrm C_{c}$ 
 be the transform taking points from the coordinate system of camera $c$ to the coordinate system of the reference camera $c^*$, and  let
 \begin{equation}
     \mathscr{C}=\{\mathrm C_{c}, c\in \mathcal{Q}\},
 \end{equation}
 be the set of transforms moving points from every camera in $\mathcal{Q}$ to the reference camera $c^*$. Indeed  $\mathrm C_{c^*}$ is the $4\times 4$ identity matrix. 
 
 Likewise, assuming that markers remain fixed with respect to each other, let $\mathrm M_{m}$ be the transform taking points from the coordinate system of  marker $m$  to the coordinate system of marker $m^*$, and let 
 \begin{equation}
     \mathscr{M}=\{\mathrm M_{m}, m\in \mathcal{M}\},
 \end{equation}
 be the set of transforms from all other markers to the reference one. Indeed  $\mathrm M_{m^*}$ is the $4\times 4$ identity matrix.

Along the video sequence,  each camera captures a total of $r$ images synchronously with the rest of cameras.  Then, let $\mathrm G_{t}$ be the transform from the marker reference system $m^*$ to the camera reference system $c^*$  at time $t$, and let 
\begin{equation}
     \mathscr{G}=\{\mathrm G_{t}, t\in\{1,\ldots,r\}\},
 \end{equation} 
be the set of all transforms along the video sequence.

Our goal is to estimate $\mathscr{C}$, $\mathscr{M}$ and $\mathscr{G}$ from the input video sequences. To do so, the images are automatically analyzed in order to find the markers corners. Then, the input of our algorithm is the set of 2D projections of the squared markers in the images where they were detected.

Let $p^{t,l}_{c,m}$ be the  image coordinates  of the  corner $u^l_m$ at time $t$ in the camera $c$. We keep all possible combinations of such indices ($t,c$ and $m$) as tuples in the set $\Upsilon$. In other words, $(t,c,m)\in\Upsilon$ means that marker $m$ is detected in camera $c$ at  time $t$.

Our solution to estimate the unknowns consists in minimizing the reprojection error of the observed markers in all cameras along the video sequence, that is defined as  \begin{equation}
     \mathrm E(\mathscr{C},\mathscr{M},\mathscr{G}) = \sum_{(t,c,m)\in\Upsilon}
     \sum_{l=1}^{4}  \left(\Psi(\mathrm C_c^{-1} \mathrm  G_t \mathrm  M_m,\theta_c) - p^{t,l}_{c,m}\right)^2.
     \label{eq::global_reprj_err}
 \end{equation}
 
Here,  $\Psi:\mathbb{R}^3 \xrightarrow{} \mathbb{R}^2$ is the perspective projection function which indicates the pixel coordinates on which a 3D point projects on a camera with intrinsic parameters $\theta_c$, and $()^{-1}$ denotes the inverse transform. 

Equation~\ref{eq::global_reprj_err} is a  non-linear function that can be minimized using the Levenberg-Marquardt algorithm  \cite{IMM2004-03215}. However, it is a local optimization method requiring an initial estimate of the unknowns. To find the initial estimation, we operate in the following manner. First, the relative position of the cameras $\mathscr{C}$ is
estimated based on the detected markers. Then, using $\mathscr{C}$, the relative marker positions $\mathscr{M}$ can be estimated along with $\mathscr{G}$. Once the initial estimations are obtained, Equation~\ref{eq::global_reprj_err} is minimized in order to globally reduce the reprojection error in the whole video sequence.

In the following sections, we provide a detailed description of how the initial estimations of $\mathscr{C}$, $\mathscr{M}$ and $\mathscr{G}$ are obtained.

\subsection{Initial estimation of   $\mathscr{C}$ }
The estimation of the relative transform between the cameras and reference camera $c^*$ is done by computing first the pair-wise relationships between cameras. Whenever two cameras observe a marker at the same time, it is possible to obtain a relative pair-wise transform between the cameras. However, because of the ambiguity problem, it is sometimes impossible to know if the transform is reliable. By aggregating such transforms over the sequence, we can discard outliers and obtain the best-observed configuration. Then, a graph representing the pair-wise relations is explored to find the optimal path between each camera and the reference one. Below, we provide a detailed explanation of the proposed method.
 
 \begin{figure}[t]
    \centering
     \includegraphics[width=0.35\textwidth]{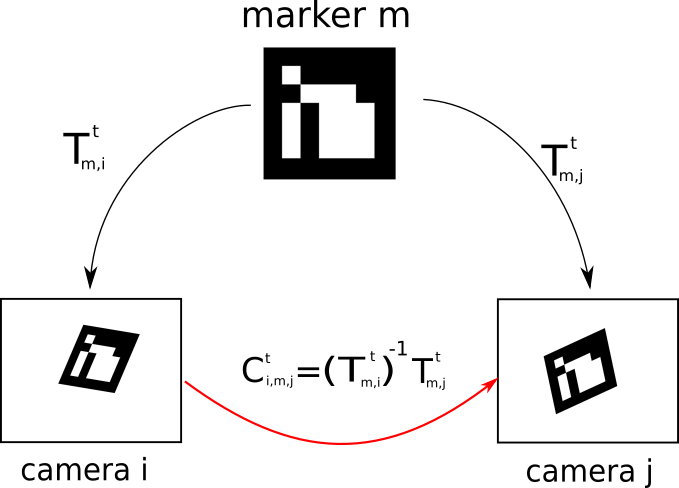}
    \caption{ Estimation of the the pair-wise camera transform $\mathrm C_{i,m,j}^t$ between cameras $i$ and $j$ from   time $t$ using marker $m$. }
     \label{fig::camerapairwise}
\end{figure}

\subsubsection{Pair-wise relationships}
Let us consider that marker $m$ is observed in camera $c$ at time $t$.  Let us denote $\mathrm{T}^t_{m,c}$ the transform   from the marker $m$ to the camera $c$ at time $t$, which  can be obtained using a planar pose estimator such as \cite{schweighofer_robust_2006}. However, the estimation of the pose using four co-planar corners is subject to ambiguity;  under some circumstances, such as errors in the estimation of the corners or low image resolution, two different poses $\mathrm{T}^t_{m,c}$ and $\dot{\mathrm{T}}^t_{m,c}$ may be valid. Nevertheless, it is possible to detect when such circumstance happens. As already indicated, the robust planar pose estimation method \cite{Collins2014} provides the two poses that explain the pose of an observed marker w.r.t to the camera. Then, it is possible to analyze the reprojection error ratio of the two solutions:
\begin{equation}
 r^t_{m,c}=\frac{  e(\dot{\mathrm{T}}^t_{m,i}) }{e(\mathrm T^t_{m,i})} \in [1,\infty),
 \label{eq::error_ratio}
\end{equation}
\noindent and use this value as a confidence measure. When it $r^t_{m,c}$, it is difficult to determine which of the two solutions is the correct one. As $r^t_{m,c}$ increases, so does the confidence in $\mathrm T^t_{m,i}$ to be correct solution.

Then, we shall denote  $\Xi^t_{m,c}$ to the set  of possible transforms from the marker $m$ to  camera $c$  at time $t$:

  \begin{equation}
   \Xi^t_{m,c} = \begin{cases}
   
   \{\mathrm{T}^t_{m,c}\} &\quad(t,c,m)\in\Upsilon~\wedge ~r^t_{m,c} \geq \tau_e \\
   
   \{\mathrm{T}^t_{m,c},\dot{\mathrm{T}}^t_{m,c}\} &\quad(t,c,m)\in\Upsilon~\wedge~r^t_{m,c} < \tau_e\\

   \emptyset &\quad\mathrm{otherwise}
  \end{cases}
  \label{eq::Xicases}
 \end{equation}

Let us now consider that the same marker $m$ is observed in two cameras $i$ and $j$ at time $t$. Then, it is possible to estimate the relative pose between the cameras as

 \begin{equation}
\mathrm C_{i,m,j}^t =\Phi(\mathrm T^t_{m,i},\mathrm T^t_{m,j})= (\mathrm T^t_{m,i})^{-1} \mathrm T^t_{m,j},
 \end{equation}
 as shown in Figure~\ref{fig::camerapairwise}.

  Since the sets $\Xi^t_{m,i}$ and $\Xi^t_{m,j}$ can have more than one transformation, we shall define the set of all possible combinations of transformations as
  \begin{equation}
  \Lambda_{m,i,j}^t =  \Xi^{t}_{m,i}\times \Xi^{t}_{m,j}.
  \end{equation}
  where $\times$ is the Cartesian product operator. We shall denote
  
  \begin{equation}
  \mathscr{C}^t_{m,i,j}=\{\Phi(\mathrm{C})~|~\mathrm{C}\in \Lambda_{m,i,j}^t\},
  \end{equation}
  to the set of possible pair-wise relationships between the two cameras given $\Lambda_{m,i,j}^t$. Even more, since the cameras can observe several markers at the same time along the sequence, let 
  \begin{equation}
  \mathscr{C}_{i,j}= \bigcup_{m=1}^{q}\bigcup_{t=1}^{r} \mathscr{C}^t_{m,i,j}=\{ \mathrm{C}_{k,i,j},k\in\{1,\ldots,s_{i,j}\}\}.
  \end{equation}
  be the set of  $s_{i,j}$  pair-wise relationships found between cameras $i$ and $j$ in the video sequence.

  \subsubsection{Optimal transforms}
\label{subsec::opttransform}

  Ideally, since the cameras do not move with respect to each other, the transforms in $\mathscr{C}_{i,j}$ should all be identical. However, due to noise and to the ambiguity problem (Sect~\ref{sec::ambiguity_problem}), this is not true. So, we need to estimate the optimal transform between the cameras $\widetilde{\mathrm C}_{i,j}$ among the elements of the set. 
  Averaging the values is not a good idea because the set $\mathscr{C}_{i,j}$ may contain many outliers.

 

  
  Our approach then is to select the transform that better explains the observations by using an indirect method that minimizes the distance of one transform to the rest of transforms. 
  
  Let us consider three arbitrary points 
  $$(v^0,v^1,v^2)~|~v^i\in\mathbb{R}^3~\wedge~v^i\ne(0,0,0)$$
   that $\mathrm C_{k,i,j}\in \mathscr{C}_{i,j}$ transform to
    $(v^0_{k,i,j},v^1_{k,i,j},v^2_{k,i,j})$. Then,     we define   
   \begin{equation}
    \mathrm d(\mathrm C_{k,i,j})= \sum_{s=1}^{s_{i,j}} \sum_{l=1}^3 \|v^l_{k,i,j}-v^l_{s,i,j}\|^2_2
    \label{eq::dcam}
   \end{equation}
   as the sum of distances from the points transformed by $C_{k,i,j}$ to the points transformed by all the elements in $\mathscr{C}_{i,j}$. Our approach is to consider the transform that minimizes the sum of this distance:

  \begin{equation}
   \widetilde{\mathrm C}_{i,j} = \argmin_{\mathrm{C}\in\mathscr{C}_{i,j} }  \mathrm d(\mathrm C)
  \end{equation}

   \begin{figure}[t]
    \centering
     \includegraphics[width=0.35\textwidth]{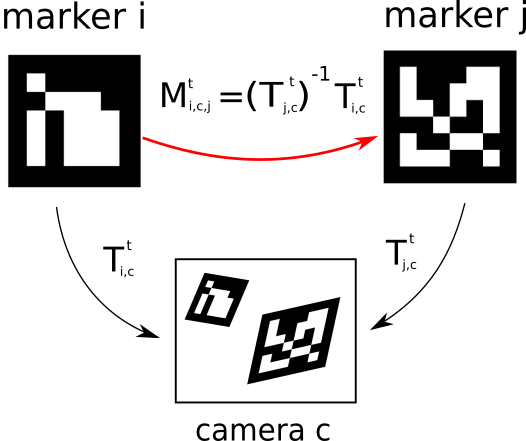}
    \caption{   Pair-wise marker transform between markers $i$ and $j$ from camera $c$ at time $t$.  }
     \label{fig::markerpairwise}
\end{figure}
  \subsubsection{Graph analysis}
\label{subsec::graphana}
  So far, we have obtained pair-wise relationships between the cameras. Our final goal in this Section is to obtain the initial estimations $\{\mathrm C_c, c\in \mathrm Q\}$ for the relative transform from each camera $c$ to the reference camera $c^*$. To do so, we are creating a graph representing the camera configuration and a minimum spanning tree is used to find the optimal path from each camera to the reference one.
  
  In our graph, vertices represent the cameras while  edges represent the quality of the pair-wise estimations. A pair-wise estimation is considered reliable if the distance $ \mathrm d(\mathrm C_{k,i,j})$ is small, and it is computed from a large number of observations $s_{i,j}$. Thus, the edge $e_{i,j}$ between vertices $i$ and $j$ is computed as:
  \begin{equation}
    \begin{aligned}
  e_{i,j} = \bar{d}(\widetilde{\mathrm C}_{i,j}) w_{i,j}\\ 
  \bar{d}(\widetilde{\mathrm C}_{i,j})=\frac{\mathrm d(\widetilde{\mathrm C}_{i,j})}{s_{i,j}}\\
  w_{i,j}=max\left(1,\frac{\tau_n}{s_{i,j}}\right)
    \end{aligned}
 \label{eq::norm_dist_graph}
  \end{equation}
  The value $e_{i,j}$ is the average distance $ \bar{d}(\widetilde{\mathrm C}_{i,j})$ affected by a weighting factor $w_{i,j}=[1,\infty)$  that accounts for the number of elements in $\mathscr{C}_{i,j}$, i.e.,  $s_{i,j}=|\mathscr{C}_{i,j}|$. The basic idea is that a camera transform obtained with few markers (less than $\tau_n$) is not very reliable, and thus  $e_{i,j}$ is increased to discourage this graph edge from being selected. When  $s_{i,j}$ is above $\tau_n$, $w_{i,j}=1$, thus not affecting the distance value. 
  
  The estimation of the marker configuration can be done using a similar strategy than for the cameras. Whenever two markers are seen in the same camera, it is possible to obtain an estimation of their relative position. Thus, for each pair of makers, we obtain the observed relative transforms along the video sequence and then select the optimal one as previously explained. Finally, we construct the graph and proceed as in Sect.~\ref{subsec::graphana}.

  To obtain the initial camera estimation, we find the minimum spanning tree on the graph and use the path on the tree from each camera $c$ to the reference camera $c^*$  as:
\begin{equation}
  \mathrm C_c = \widetilde{\mathrm C}_{c^*,c_1}\widetilde{\mathrm C}_{c_1,c_2}\ldots\widetilde{\mathrm C}_{c_{m-1},c_m}\widetilde{\mathrm C}_{c_m,c}.
\end{equation}
\begin{figure}[t]
    \centering
     \includegraphics[width=0.35\textwidth]{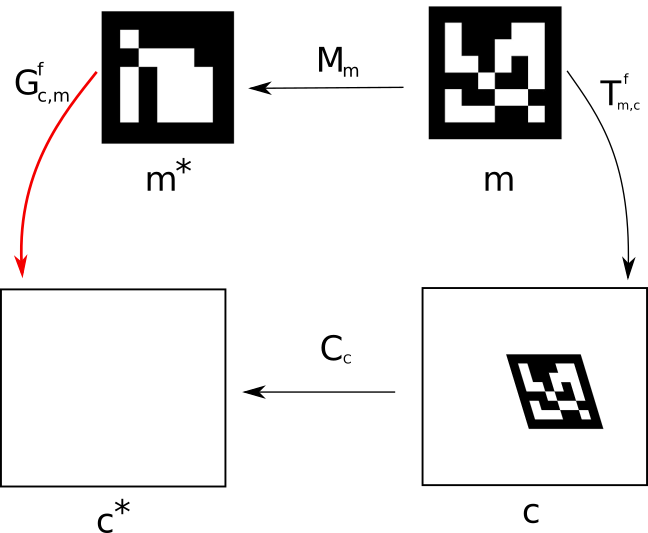}
    \caption{ Estimation of  $\mathrm{G}^t_{c,m}$ from a marker $m$ observed in camera $c$ at time $t$}
     \label{fig::glocalcaminframe}
\end{figure}

\subsection{Initial estimation of   $\mathscr{M}$}

Let us consider  two different markers $i$ and $j$ observed in a camera $c$, whose relative transformation from the markers to the camera are given by $\mathrm{T}^t_{i,c}$ and $\mathrm{T}^t_{j,c}$. Then, the transformation from marker $i$ to marker $j$ can be obtained as:
\begin{equation}
  \mathrm{M}^t_{i,c,j}= \Phi( \mathrm{T}^t_{i,c},\mathrm{T}^t_{j,c})=\left(\mathrm{T}^t_{j,c}\right)^{-1}\mathrm{T}^t_{i,c},
\end{equation}
as shown in Figure~\ref{fig::markerpairwise}.

As previously indicated, we can have more than one valid transformation per marker in a single image due to the ambiguity problem. Thus, we defined $\Xi^t_{i,c}$ and $\Xi^t_{j,c}$ as the set of transformations from marker $i$ and $j$ to camera $c$, respectively. Let then 
\begin{equation}
 \Theta_{i,j}=\bigcup_{c=1}^{n}\bigcup_{t=1}^{r}\Xi^t_{i,c}\times\Xi^t_{j,c}, 
\end{equation}
be the set of combinations of such transformations,
\begin{equation}
\mathscr{M}_{i,c,j}^t=\{\Phi(\mathrm{M})~|~\mathrm{M}\in\Theta_{i,j}\},
\end{equation}
the set of all pair-wise relationships between the markers $i$ and $j$ at time $t$ according to camera $c$, and
\begin{equation}
\mathscr{M}_{i,j}=\bigcup_{c=1}^{n}\bigcup_{t=1}^{r}  \mathscr{M}_{i,c,j}^t,
\end{equation}
the set of all pair-wise relationships between the markers $i$ and $j$ collected along the video sequence in all the cameras.

We apply the same rationale in Sect.~\ref{subsec::opttransform} to find the optimal transformations $\widetilde{ \mathrm{M}}_{i,j}\in\mathscr{M}_{i,j}$, and then the graph approach in Sect.~\ref{subsec::graphana} to obtain the best relative transformation $ \mathrm{M}_m$ of marker $m$ to the reference marker $m^*$.

\subsection{Initial estimation of $\mathscr{G}$}


  \begin{figure*}
    \centering
    \subcaptionbox{\label{fig:overall_setup}}{
        \includegraphics[height=4.cm]{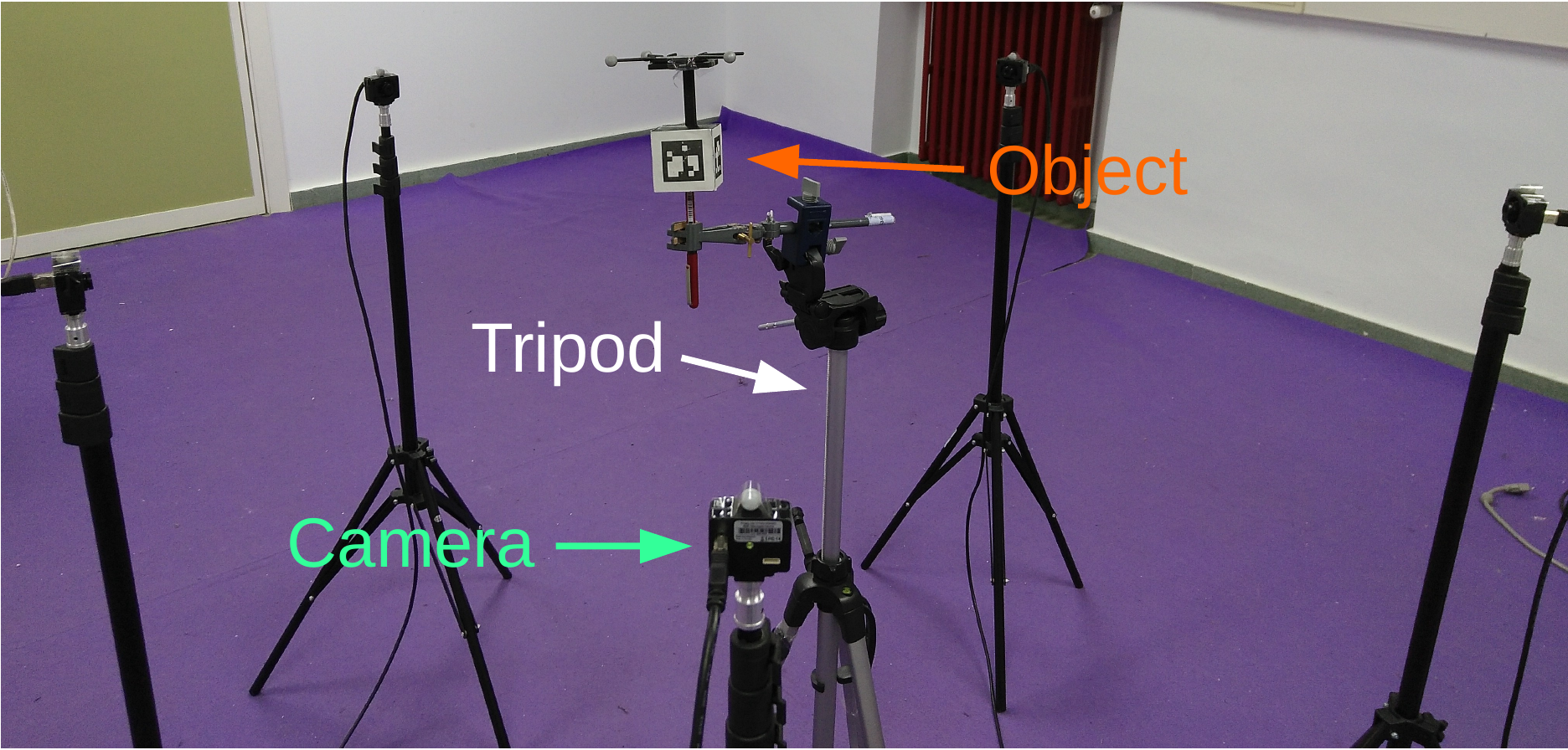}
    }
    \subcaptionbox{\label{fig:close_up_setup}}{
        \includegraphics[height=4.cm]{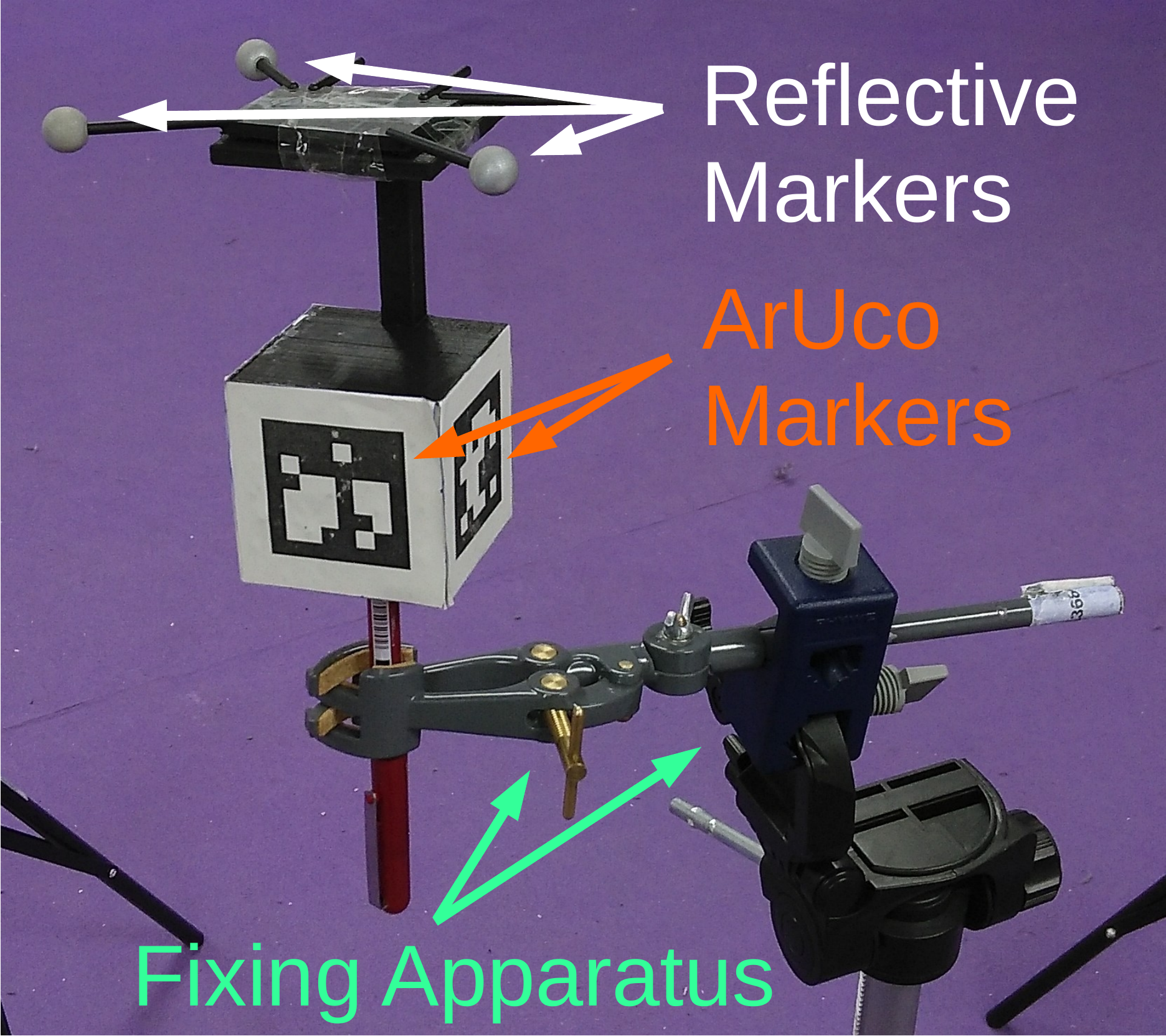}    
    }
    \subcaptionbox{\label{fig:close_up_cam}}{
        \includegraphics[height=4.cm]{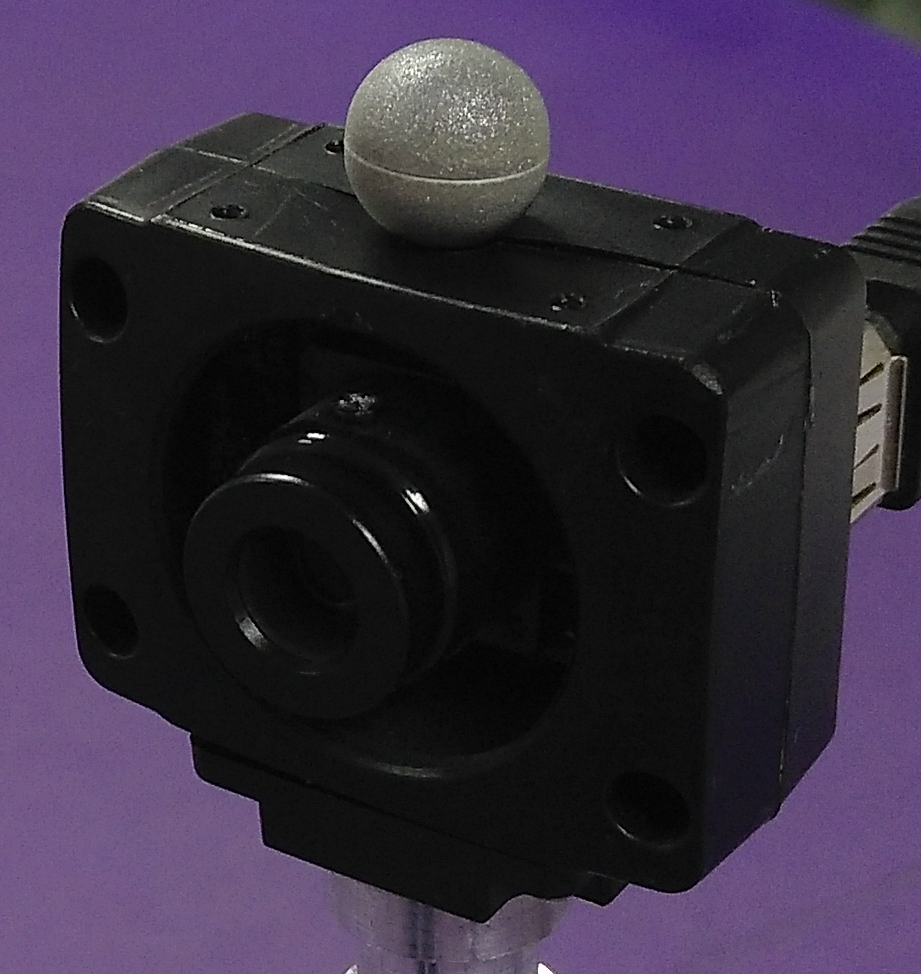}    
    }
    \caption{An illustration of (\subref{fig:overall_setup}) our overall evaluation setup and (\subref{fig:close_up_setup}) a close-up of the object being employed with the annotation of the reflective markers used by the motion capture system to track the object, the ArUco markers that are used by our system, and the apparatus used to fixes the object in its place and (\subref{fig:close_up_cam}) and example of the position of the reflective marker on the camera. The marker is used for evaluating the estimation of camera position.}
\end{figure*}
At this point, an estimation of the cameras' structure $\mathscr{C}$ and of the markers' structure $\mathscr{M}$ has been obtained. Our goal now is to estimate the set $\mathscr{G}$, that relates the set of markers with the reference camera at time $t$. To do so, we only consider the relative transform from the reference marker $m^*$ to the reference camera $c^*$, since the transform from all markers to the reference marker is already known.

Every detection of a  marker $m$ in a camera $c$ at time $t$ produces an estimation of the relative pose between the reference marker and the reference camera given by:
\begin{equation}
 \mathrm{G}^t_{c,m}=  \mathrm{C}_c   \mathrm{T}^t_{m,c} \mathrm{M}_m^{-1},
 \end{equation}
as shown in Figure~\ref{fig::glocalcaminframe}. Consequently,  let
\begin{equation}
  \Pi^t = \bigcup_{(t,c,m)\in\Upsilon(t)} \mathrm{G}^t_{c,m}
  \end{equation}
  be the set of transformations from the reference marker to the reference camera given all detected markers, in all cameras, at time $t$, where $\Upsilon(t)$
  is the set of all detected markers in all cameras at time $t$.

  As in the previous occasions, we must select the best transformation $\mathrm{G}^t\in\Pi^t$. To do so, we apply the same rationale as in Sect.~\ref{subsec::opttransform}. In other words, the one that minimizes the distance to all the others.

 \subsection{Global Optimization}
 Once initial estimates of  $\mathscr{C}$,  $\mathscr{M}$ and  $\mathscr{G}$ have been obtained, it is possible to minimize Eq.~\ref{eq::global_reprj_err}. In order to speed up the computation, we do not directly optimize the $4\times 4$ transformation matrices. Instead, we extract the rotational part into three components using the Rodrigues' formula \cite{Rodrigues}, and concatenate it to the translational part, thus, representing each transformation by only six parameters. 
 
 The optimization is done  using the  Levenberg-Marquardt algorithm \cite{IMM2004-03215}, an iterative method that uses the Jacobian matrix of the error function to reduce the error. This method requires an initial estimation of the  solution $\mathbf{x_0}$, that is incrementally replaced at each iteration $\mathbf{k}$ by a new estimate 
 \begin{equation}
  \mathbf{x_{k+1}}=\mathbf{x_{k}} + \mathbf{p_{k+1}}.    
 \end{equation}

 Let $\mathbf{J}$ be the Jacobian of the error function to be minimized $f(\mathbf{x})$ (see  Eq.~\ref{eq::global_reprj_err}). The method searches at each iteration in the direction given by the solution of the equations
\begin{equation}
 (\mathbf{J_k}^\top \mathbf{J_k}+\lambda_\mathbf{k} \mathbf{I})\mathbf{p_{k+1}}=-\mathbf{J_k}^\top f(\mathbf{x_k}), 
\end{equation}

\noindent  where $\lambda_\mathbf{k}$ is a non-negative scalar and $\mathbf{I}$ is the identity matrix. The damping factor $\lambda_\mathbf{k}$  is automatically adjusted at each iteration. When the reduction of the error is large, a smaller value is employed, making the algorithm closer to the Gauss-Newton algorithm. If an iteration provides insufficient error reduction, $\lambda_\mathbf{k}$ is increased so that the method becomes more similar to the gradient descent.


Finally, please notice that in our case, the Jacobian is sparse, since in general, only a small subset of the markers project on each camera. Thus, we take advantage of sparse matrices to speed up the calculation.


 \subsection{Object tracking}
As a result of the global optimization, the configuration of the object, the camera poses, and the object poses are obtained along the frame sequence. However, global optimization is a slow process that is no longer necessary for subsequent tracking purposes (unless one of the cameras moves). Tracking can be done efficiently now since it is a process that only requires   to first detect the markers in the images, and then, to estimate the object's pose. In essence, tracking is a sub-problem of Eq.~\ref{eq::global_reprj_err}, in which only the relative pose of the reference marker $m^*$ w.r.t. the reference camera $c^*$ is estimated, i.e., $\mathrm  G_t$. Let $\Upsilon^t$ be the projections of the markers' corners in the cameras at time $t$, then, the pose $\mathrm  G_t$ is estimated as:
\begin{equation}
\mathrm E( \mathrm  G_t)= \sum_{(t,c,m)\in\Upsilon^t}
     \sum_{l=1}^{4} \left(\Psi(\mathrm C_c^{-1} \mathrm  G_t \mathrm{M}_m,\theta_c) - p^{t,l}_{c,m}\right)^2,
      \label{eq::tracking_}
\end{equation}
\noindent Again, the LM algorithm is employed for solving the equation, which this time has only six parameters to be optimized. At each time step, the previous solution $\mathrm  G_{t-1}$ is employed as starting point for optimization. For the first frame $t=0$, we must provide an initial solution. Since many markers may be visible at the same time from the cameras, we employ as starting solution the one obtaining the maximum reprojection error ratio (Eq.~\ref{eq::error_ratio}).

 %
\section{Experimental Results}

\label{sec::results}

This section explains the experiments conducted to validate the proposed method. In the first experiment (Sect.~\ref{sec::quantiative}), we perform a quantitative evaluation of the method by analyzing its accuracy in three different aspects:  the accuracy in the estimation of the pose of the object, the precision of object reconstruction, and the accuracy in estimation of camera extrinsic parameters. Then, Sect.~\ref{sec:exp:camNum} analyzes the impact of the number of cameras on the precision of the system. Afterward, Sect.~\ref{sec::exp:objconf}  analyzes the ability of the proposed method in reconstructing objects of different configurations. Finally, the test in Sect.~\ref{sec::exp:contracking}, focuses on tracking performance of the proposed method, evaluating its usefulness in applications requiring tracking of a 3D object such as a surgical tool.

For the experiments, five synchronized global-shutter cameras, with a resolution of $640\times480$ pixels, able to capture at $60$~Hz, were employed. The five cameras were placed forming a circle and pointing towards its center (see Fig~\ref{fig:overall_setup}). In addition, an Optitrack motion capture system comprised of six cameras able to track the position of reflective markers at $100$~Hz was employed to obtain the ground truth values for the object and camera poses.

Please notice that our method does not impose restrictions on the camera setup. The only requisite is that the cameras share part of the field-of-view so that the relationship between them can be established. Using exclusively the video footage showing the markers, the proposed method is able to estimate all parameters automatically. In other words, our method obtains the initial estimations of $\mathcal{C},\mathcal{M}$ and $\mathcal{S}$ automatically, and later refines them.

Along the paper, the proposed method have employed the two parameters $\tau_c$ and $\tau_n$. The first one indicates the threshold for the reprojection error ratio (Eq.~\ref{eq::Xicases}) and the second one a threshold for the graph (Eq.~\ref{eq::norm_dist_graph}). Our experience indicates that the values $\tau_c=2$ and $\tau_n=10$ are a good choice.

While doing experiments, different values for different parameters as stopping criteria of the leveberg-marquardt optimization were used. The most important parameters were minimum average error improvement in all dimensions and maximum number of iterations. For these two parameters generally the values $10^{-4}$ (pixels) and 10000 iterations yielded good results respectively.
\subsection{Quantitative evaluation of the method}
\label{sec::quantiative}
The first experiment analyzes the precision of the proposed method. Using a 3D printer, we created an object (Fig~\ref{fig:close_up_setup}) with four flat surfaces to which both square and reflective markers were attached. The rectangular markers attached had a side length of four centimeters. 

We are mainly interested in measuring the precision in the estimation of the object pose, which is an indicator of how good the system is for the tracking task. In addition, the pose of the cameras employed, along with the configuration of the employed object are also estimated by our method and its precision analyzed.

Four  video sequences placing the object at different positions of the area observed by all the cameras was recorded and also registered with the motion capture system in order to obtain the ground truth values. Reflective markers were placed both on the object (Fig~\ref{fig:overall_setup}b) and on the cameras  (Fig~\ref{fig:overall_setup}c). The object was placed in a fixing apparatus to  avoid synchronization problems between the cameras and the motion capture system.


\begin{table}
\begin{small}
\begin{center}
\begin{tabular}{l || r | r | r | r}
    Camera Circle Radius (m) & 0.7 & 0.9 & 1.1 & 1.3\\
    \hline
\hline
    Object Rotation  error (deg) & 1.11 & 1.05 & 0.88 & 1.56\\
    \hline
    Object Translation  error  (mm) & 0.68 & 0.85 & 0.97 & 0.87\\
    \hline
    Camera Translation  error (mm) & 2.14 & 2.38 & 4.18 & 5.12\\
   \hline
    Object Configuration  error (mm) & 0.21 & 0.30 & 0.55 & 0.83\\
\end{tabular}
\caption{ Errors obtained by the proposed method when placing the camera at different distances (see text for details).}
\label{error_table}
\end{center}
\end{small}
\end{table}
\begin{table*}[!tb] 
    \begin{minipage}{.48\textwidth}
      \centering
\begin{tabular}{l || r | r | r | r|r|r}
    Radius (m) & $err_1$ & $err_2$& $err_3$ & $err_4$& $err_5$ & avrg \\
    \hline
   \hline
   0.7  & 0.64&	0.73	&0.62	&0.84	&0.66&	0.70\\
    \hline
   0.9  &  0.83  & 	2.15	& 1.00 	& 1.02 & 	0.84 & 	1.17 \\
    \hline
  1.1& 1.02& 	1.13 	&1.05 & 0.95	& 0.86& 	1.00\\
    \hline
  1.3& 1.10	&0.83	&0.99&	0.95 	&0.99&	0.97\\
\end{tabular}
      \caption{Translation errors (millimetres) in the estimation of the object position using four cameras.}
      \label{tab::fourcams_translation}
    \end{minipage}%
    \hfill
    \begin{minipage}{.48\textwidth}
      \centering
      
 \begin{tabular}{l || r | r | r | r|r|r}
    Radius (m) &$err_1$ & $err_2$& $err_3$ & $err_4$& $err_5$ & avrg \\
    \hline
   \hline
   0.7  & 1.10&	1.10 & 1.10	& 1.12	& 1.06&	1.09\\
    \hline
   0.9  &  1.02  & 	1.69& 1.08 	& 1.04& 1.05& 	1.18 \\
    \hline
  1.1& 0.87& 	1.02 	&0.92& 0.90	&0.93& 	0.93\\
    \hline
  1.3& 1.57	&1.55	&1.62&	1.59 	&1.56&	1.58\\
\end{tabular}    
\caption{Rotational errors (degrees) in the estimation of the object position using four cameras.}
\label{tab::fourcams_rotation}
\end{minipage} 
\end{table*}

\begin{figure*}
    \centering
    \begin{subfigure}[h]{.45\textwidth}
        \includegraphics[height=5cm]{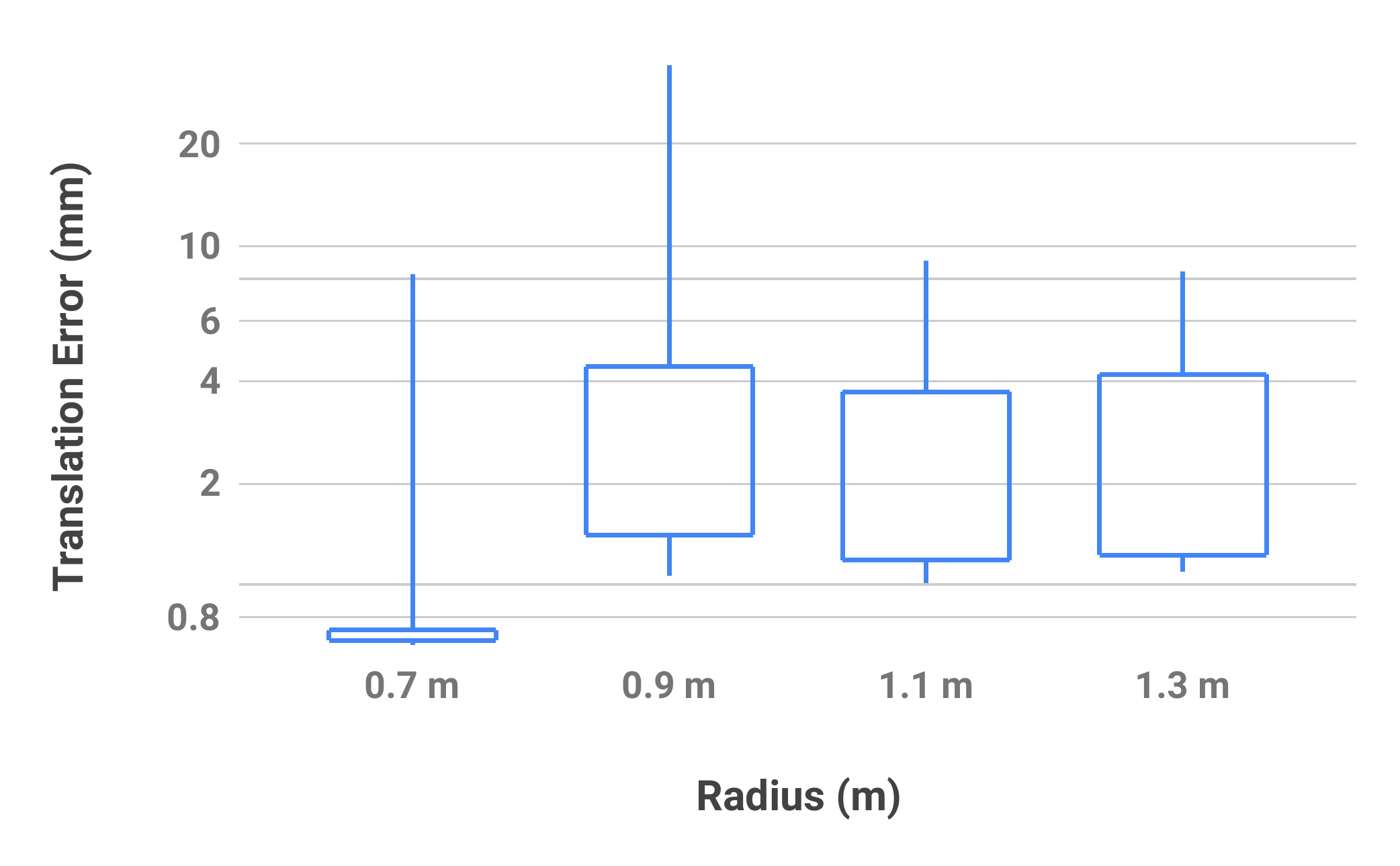}
        \caption{}
        \label{fig:error3cams:position}
    \end{subfigure}
    \begin{subfigure}[h]{.5\textwidth}
        \includegraphics[height=5cm]{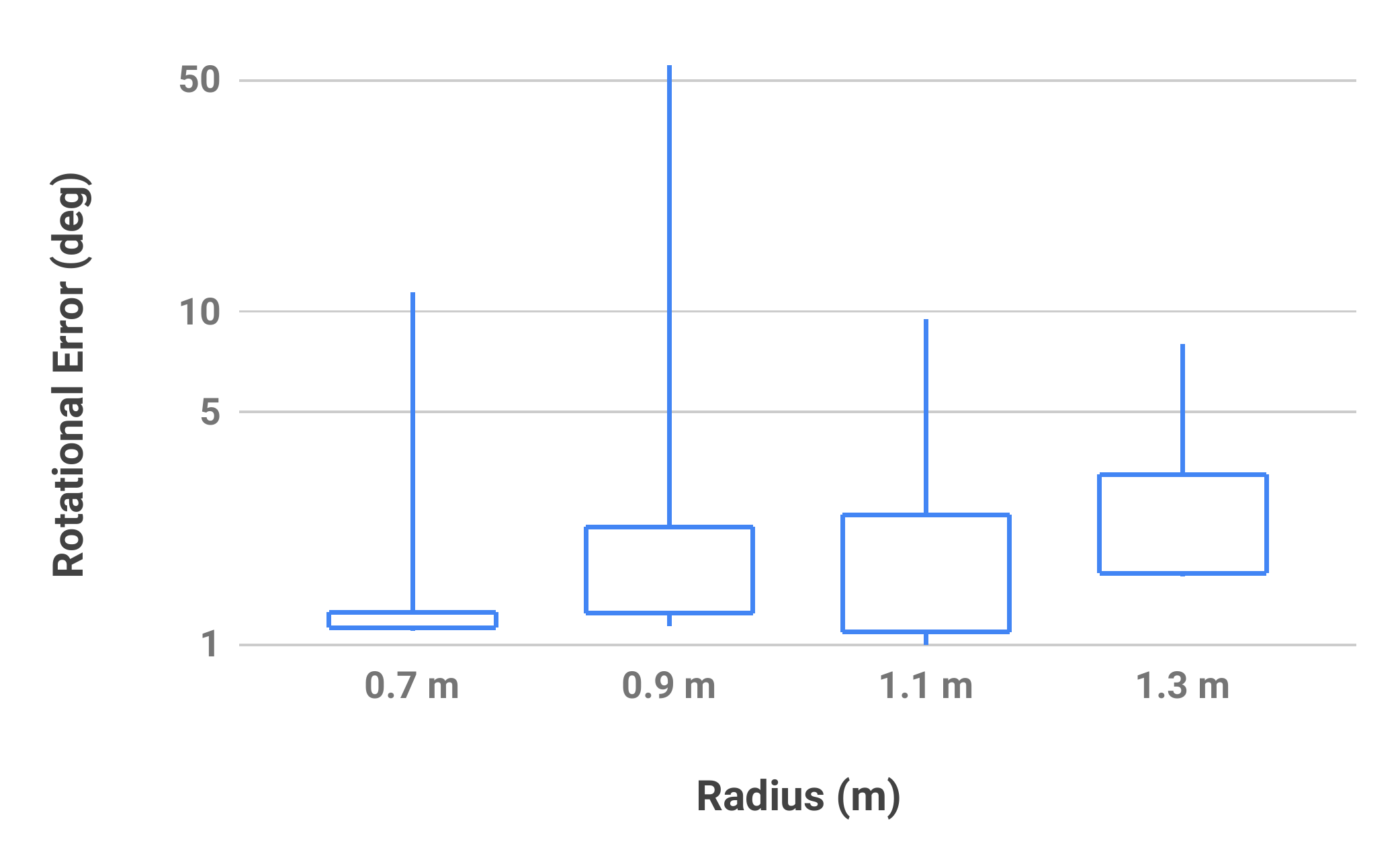}
        \caption{}
        \label{fig:error3cams:rotation}
    \end{subfigure}
    \caption{Errors obtained using three cameras. (\subref{fig:error3cams:position}) Translation errors (\subref{fig:error3cams:rotation}) Rotation Errors.}
    \label{fig:error3cams}
\end{figure*}

The precision of the system is expected to be influenced by the distance from the camera to the object, i.e., the farther the object from the cameras, the lower the accuracy. So, in order to analyze the impact of distance, we repeated the recordings at four different distances: we placed the cameras forming circles of radius $0.7,0.9,1.1$ and $1.3$ meters. These values were estimated applying the method in \cite{coope_circle_1993}.

The results obtained are shown in Table~\ref{error_table}. Columns represent the result for the different camera distances. The second and third rows indicate the error of the proposed method in estimating the pose of the object. While the second row indicates the average error in the estimation of the translation component (in millimeters), the third one indicates the average rotation error (in degrees) These errors have been computed first aligning the trajectories obtained by our approach and the Optitrack system (ground-truth), taking advantage of the Horn method~\cite{horn_closed-form_1987}.  Then the average rotation and translation distances between the trajectories are computed.

These two errors are the most important ones since they indicate the accuracy of the proposed method for tracking tasks, e.g.,  augmented reality applications or in surgical tool tracking. It can be observed that, in most cases, the proposed system achieves sub-millimeter accuracy.  Please notice that we are employing low-resolution cameras, thus limiting the maximum distance from the object to the marker. Using cameras of a higher resolution would allow increasing the working distance without compromising accuracy, at least in theory.

The fourth row shows the error in estimating the position of the camera. As previously indicated, we placed a reflective marker at the top of each camera. Therefore, we  obtained the relative position of the cameras and compare it with the poses estimated by our method. The positions provided by the Optitrack system  where aligned to the 3D positions obtained by our method (taking advantage of the Horn algorithm~\cite{horn_closed-form_1987}) and the average distance between both is considered the error. It can be observed that the errors are slightly higher. However, since the placement of the reflective markers was manual, it is subject to more errors and thus it is a less reliable measure. 

The fifth row of Table~\ref{error_table} shows the error that our system obtains in estimating object configuration. In particular, it refers to the relative positions of the squared markers. To evaluate the error, we took high-resolution pictures of the 3D object created and process them with the method  \cite{munoz-salinas_mapping_2018}, which is able to provide the relative pose of the markers. Assuming it is the ground truth method, we obtained errors that, in the worst case, are around one millimeter.

Finally, let us indicate that on average, the optimization step took around five minutes in a core $i5$ laptop computer running Ubuntu 16.04.  

\subsection{Influence of the number of cameras}
 \label{sec:exp:camNum} 
In order to test the robustness of the method to the number of cameras, we processed the sequences excluding some of them from the processing. Tables~\ref{tab::fourcams_translation} and \ref{tab::fourcams_rotation} show the translation and rotation errors obtained when the videos previously employed are processed using four of the five available cameras. All possible combinations have been tested and reported in the table under the column $err_1\ldots err_5$.  In the tables, the column label $err_i$ means that the view of camera $i$ has not been used. As can be observed, the behavior of the system using four cameras is very similar to as observed before.  The final column $avrg$ represents the average error obtained. 

The same experiments were done using only three of the five cameras available. In that case, the number of possible combinations is higher and thus the results are shown in Figs~\ref{fig:error3cams}(a,b). One can observe cases in which the errors are very high. This cases, outliers, occurs because of the initialization was not able to provide a solution good enough for the global optimization to start from a promising location. As a consequence, the method is trapped in a local minimum. Nevertheless, in general, the proposed system is able to achieve good solutions in most of the cases. In particular, the median translation errors (in millimetres) are $[0.70,2.02,1.24,1.29]$ for the camera distances analyzed, and the median rotational errors (in degrees) are $[0.91,1.07,0.88,1.05]$.

Since we were interested in having a full 360 degree view of the object we did not test our algorithm using less than three cameras.

\subsection{Testing different object configurations}
\label{sec::exp:objconf}
Experiments were conducted to show that the proposed method is able to employ different type of objects comprised by several planar markers. In particular, we have tested our method with the three objects shown in Figure~\ref{fig::teaser}(b).

As in the previous case, the five cameras were employed to record a video sequence in which each object was freely moved. Then, our method was applied and the object configuration obtained compared with the one obtained by the method in \cite{munoz-salinas_mapping_2018}.

\begin{figure*}[t] 
     \includegraphics[width=\textwidth]{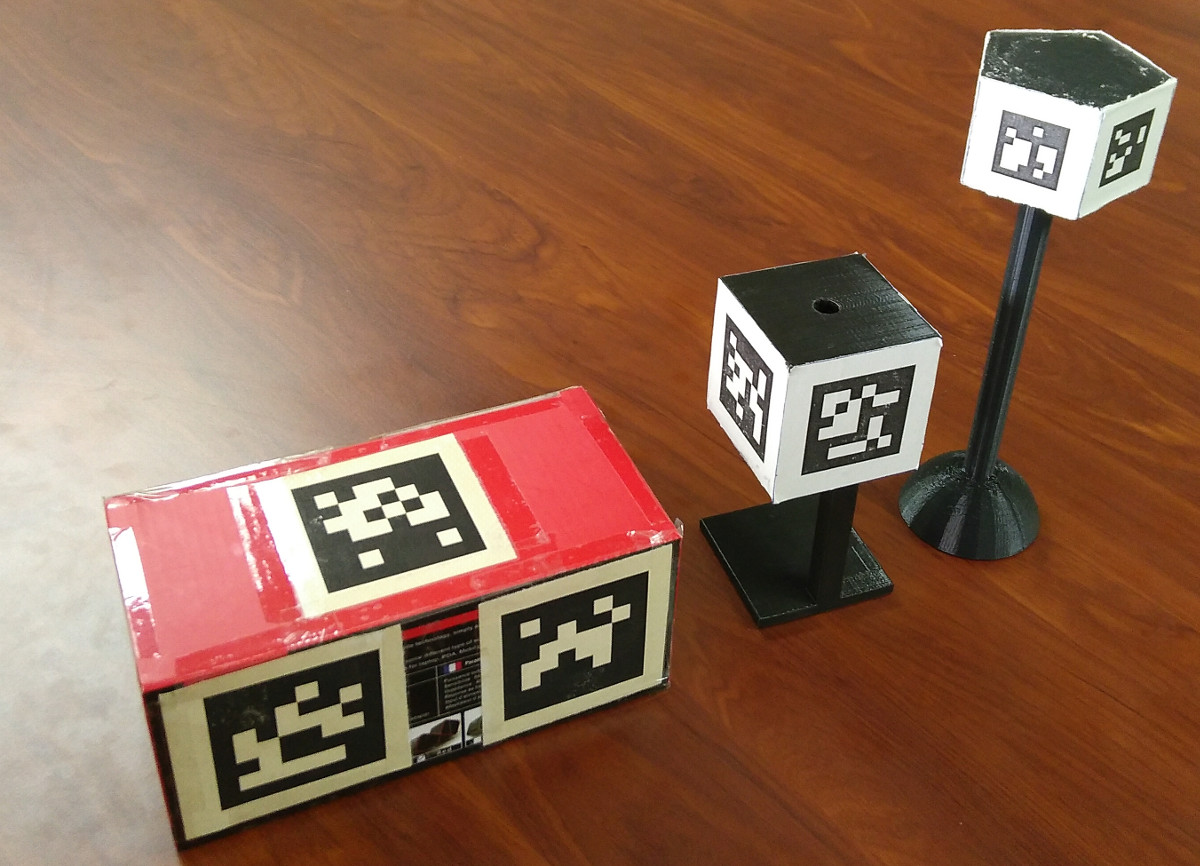}
    \caption{ Three different objects employed in our tests shown in Fig~\ref{fig::teaser}  before and after optimization. (a) Four sided object (b) Box with markers attached in random locations. (c) Pentagon object.}
     \label{fig:objetcs}
\end{figure*}

Figure \ref{fig:objetcs} shows the 3D reconstructions obtained by our method, before and after optimization. In the case of the four sided object (Fig.~\ref{fig:objetcs}a), the initial reconstruction is very good and optimization provides little improvement. In the second case (Fig.~\ref{fig:objetcs}b), the initial object reconstruction is not very exact and it can be observed that the markers are not in the same planes. However, after the optimization the reconstructed object is corrected. Finally, the pentagon (Fig.~\ref{fig:objetcs}c) is an object with much smaller markers, that occupy a small portion of the visible images. As can be observed, the initial reconstruction is clearly wrong, showing one of the markers out of its place. Nevertheless, the optimization method is able to properly correct the object.

When the objects reconstructed with our method are compared to \cite{munoz-salinas_mapping_2018} as the ground truth, the errors obtained are $[0.2,0.5,0.4]$~millimetres. However, please notice that the images employed for ground truth are high resolution images ($3200\times2400$), while the images employed by our method have a resolution of only $640\times 480$.

\subsection{Continuous tracking}
\label{sec::exp:contracking}

\begin{figure*}[!ht]
    \centering 
    \begin{subfigure}[h]{.45\textwidth}
        \includegraphics[height=5cm]{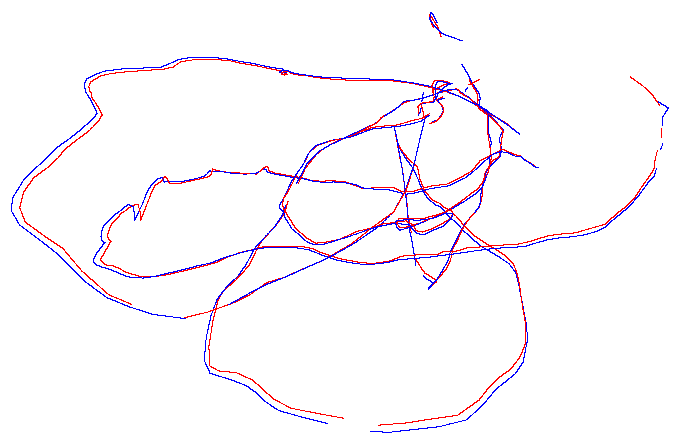}
        \caption{}
        \label{fig:trajectories}
    \end{subfigure}
    \begin{subfigure}[h]{.54\textwidth}
        \includegraphics[height=5cm]{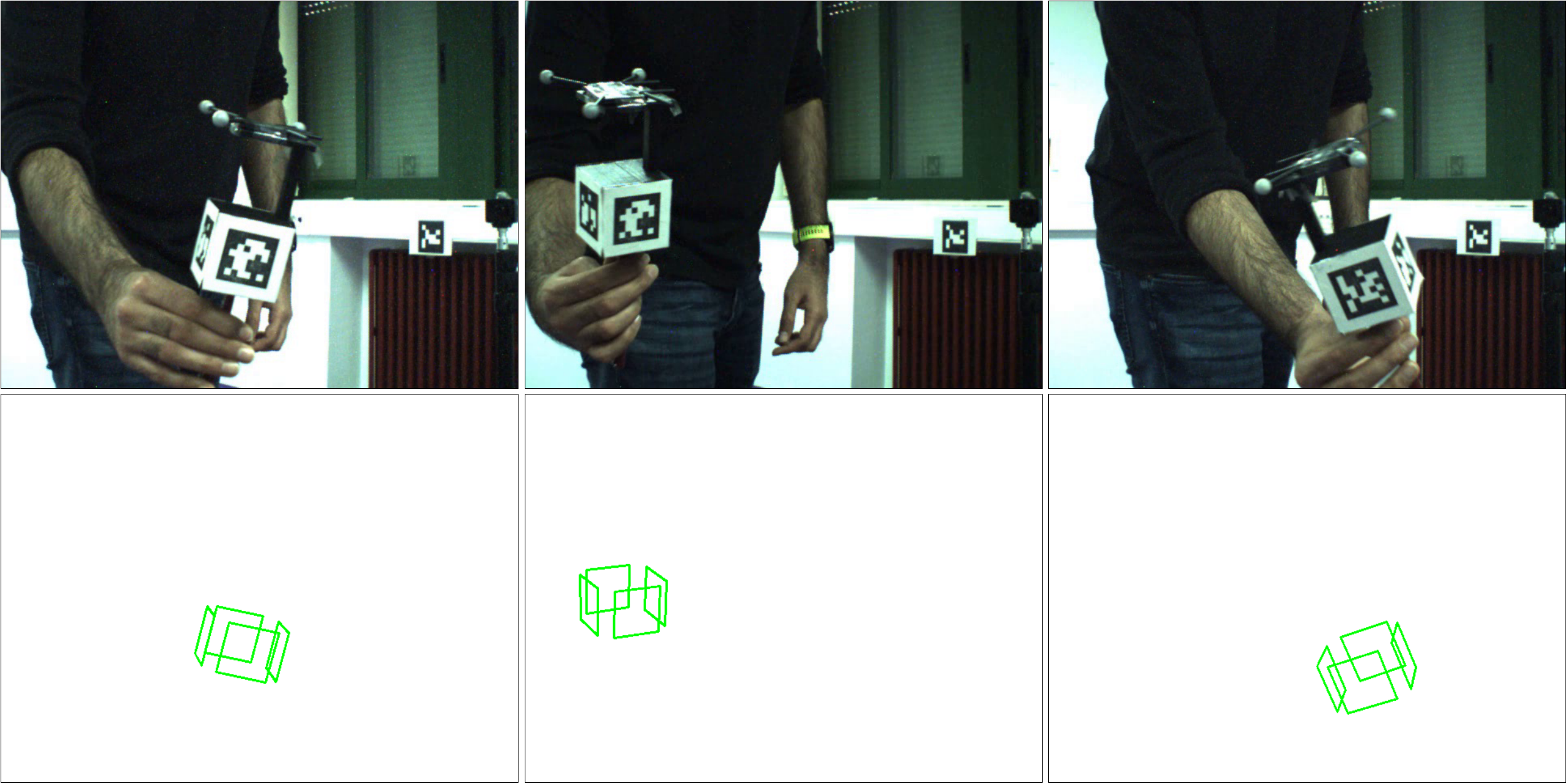}
        \caption{}
        \label{fig:handling}
    \end{subfigure}
    \caption{(\subref{fig:trajectories}) Object trajectories of the ground truth from the motion capture system (the red curve) and our method (the blue curve) using RGB cameras aligned after finding the transformation between their two coordinate system and the local transformation between the reflective markers and the reference ArUco marker. (\subref{fig:handling}) different frames from the same sequence showing the original frames from one of the cameras (top) and the corresponding results from our algorithm (bottom).}
    \label{fig:tracking}
\end{figure*}

Finally, we tested the capability of our method for the tracking task. In a realistic scenario, the full pipeline of our method is only used once in order to obtain the parameters of cameras and object. However, once they are known, tracking can be done at a higher speed. Figure~\ref{fig:tracking} shows the tracking results of our method in one of the sequences recorded.  Fig.~\ref{fig:tracking}(a) depicts in red color the real trajectory of the object, according to the motion capture system, while in green it is the trajectory estimated by our method.  This is challenging case where the object was moved at high speed at some parts of the sequence, causing blurring and synchronization problems with the motion capture system. It can be observed that in some parts of the trajectory the estimated position differs more from the ground truth than in others, because of the higher speed. In some other parts, due to occlusions, it is not possible to do tracking. In this sequence, the tracking error is of $5$ millimeters.

Figure~\ref{fig:tracking}(b) shows three snapshots of the video sequence. Images at the top are captured by the camera, while the bottom images show the estimated three-dimensional object poses.


\subsubsection*{Tracking Speed}
In order have a better understanding of the computational demands of the proposed method, it has been evaluated in a range of situations. Table \ref{tab:computing_times} reports the frame rate of the proposed method for tracking, considering that the camera and object configurations have been obtained. Experiments were done for different image resolutions and number of threads. Since the videos have been recorded using cameras of $640\times480$ pixels, the images have been upsampled and downsampled to obtain different resolutions. The results reported are obtained over a sequence of $735$ frames, using different numbers of CPU threads and parallelization with the OpenMP API. The test was done on an Intel Core i7 machine with four computing cores plus hyper-threading. As you can see, the best performances on the machine are achieved using five or six threads.

\begin{table}[]
    \centering
    \begin{tabular}{c||m{1.15cm}|m{1.15cm}|m{1.3cm}|m{1.45cm}}
         \multirow{2}{1.25cm}{\centering Number of Threads} & \multicolumn{4}{c}{Frame Resolution} \\
           & $320\times240$& $640\times480$ & $1280\times960$ & $1920\times1440$\\
         \hline \hline
          1 & 234 & 128 & 68 & 38\\
          \hline
          2 &  347 &  167&  83& 47\\
          \hline
          3 &  356 &  200&  91& 48\\
          \hline
          4 &  344 &  210 &  90 & 49\\
          \hline
          5 &  445 &  \textbf{237}&  \textbf{96}& \textbf{51}\\
          \hline
          6 &  \textbf{449}&  223&  92& \textbf{51}\\
          \hline
          7 &  361&  199&  91& 47\\
          \hline
          8 &  314&  172&  45& 22\\
    \end{tabular}
    \caption{Tracking speed (frames per second)  for different frame resolutions and number of threads.}
    \label{tab:computing_times}
\end{table}
Tracking consists in detecting markers in the images and then estimating the camera pose. While the computing time employed in marker detection depends on the image resolution, object pose estimation in a sequential process consumes approximately $1.5$ milliseconds per frame in our tests.

\subsection{Comparison to other methods}

\subsubsection*{Extrinsic Camera Calibration}

This section analyzes the accuracy of our method for  extrinsic camera calibration and compares it with other available methods for the same task using the same camera setup employed in previous experiments.

Three different calibration approaches available in the OpenCV library were evaluated: the commonly used chessboard pattern, the asymmetric circle grid pattern, and a multi-camera calibration tool which uses a random pattern \cite{li_multiple-camera_2013}. 
The first two methods are generally employed for single camera calibration. We adapted them for multiple cameras by  first estimating the  pairwise extrinsic parameters of adjacent cameras (using the {\it stereoCalibrate} function from the OpenCV library \cite{bradski_opencv_2000}). Then a root camera was chosen and the extrinsics of the other cameras were calculated with respect to it. Finally, the optimal rigid transformation and scaling was estimated~\cite{horn_closed-form_1987} to match the camera center positions to those that were measured by the motion capture system. For the multi-camera calibration tool, we recorded images and run the method as indicated in the documentation. It is important to remark that this method internally estimates the camera intrinsics too.

The  errors obtained for the different camera configuration are shown in Table~\ref{tab:calib_errors}. The first column of the table shows the results of our approach, already reported in Table~\ref{error_table}. They have been set out again to ease the comparison.
\begin{table}[]

    \centering
    \begin{tabular}{c||c|c|c|c}
         \multirow{3}{1.5cm}{Radius(m)} & \multirow{3}{1.3cm}{Our Approach} & \multirow{3}{1.6cm}{Chessboard} & \multirow{3}{1.7cm}{Asymmetric Circle Grid} & \multirow{3}{1.2cm}{Multi-Camera \cite{li_multiple-camera_2013}}  \\
         & & & & \\
         & & & & \\
       \hline \hline
         0.7 & 1.68 & 2.36 & 5.65 & 122.41\\
        \hline
         0.9 & 2.41 & 2.54 & 3.90 & 75.76\\
        \hline
         1.1 & 4.90 & 2.80 & 7.60 & 65.80 \\
         \hline
         1.3 & 3.73 & 6.68 & 16.86 & 795.04
    \end{tabular}
    \caption{Camera translation error (in millimeters) using different extrinsic calibration methods at different distances.}
    \label{tab:calib_errors}
\end{table}
 
As can be noticed, our method obtains the best results in almost all cases. The exception is at distance $1.1$ meters, in which the ChessBoard method obtains better results. The Assymetric Circle Grid method performs worse than the Chessbord, and the Multi-Camera method \cite{li_multiple-camera_2013} is the worst. 

We think there are different reasons for superiority of our approach here. First, our object compared to a calibration board can be viewed by more number of cameras at the same time and optimizes all of the global extrinsics at the same time. While in the case of chessboard or asymmetric circle grid, calibrations are done pairwise and the global extrinsics are inferred from those. The second reason is that our approach optimizes the object configuration simultaneously with extrinsics calibrations while calibration boards use a theoretical 3D configuration which is not optimized. Lastly, we suspect that the attempt of the Multi-Camera approach \cite{li_multiple-camera_2013} to optimize the intrinsic parameters at the same time as extrinsics makes the problem much more difficult since we experienced the same problem while trying to implement this idea.
\subsubsection*{Object Tracking}

The method proposed in this paper is the only one in the literature, up to our knowledge, able to track multiple planar markers from multiple cameras. 

Nevertheless, in order to compare our approach with other methods, two state-of-the-art system has been selected. First, Apriltag 2 \cite{wang_apriltag_2016}, which is a method able to estimate the pose of a single marker using one camera. The second method is ArUco~\cite{Aruco2014}, which can track an object comprised of multiple markers using one camera. Since ArUco is not able to reconstruct the object configuration, MarkerMapper~\cite{munoz-salinas_mapping_2018} was employed to obtain it and provided to ArUco.

Because these methods use only one camera, we calculated their output in each camera separately and then moved the results to the global coordinates using the extrinsic calibrations obtained using the Chessboard calibration method. Since Apriltag 2 only tracks single markers (but our object has several) we compute the error of the method using (at each frame) the estimation of the detected marker that faces the camera more directly, which in general is the one providing the smallest error.

For the ArUco algorithm we take the average pose reported by each camera as the final tracking result. The results for the four different radii in our setup is reported in Table \ref{tab:tracking_errors}. As you can see, our method outperforms the others in all cases.

 We suppose the reason our approach is better than the combination of the Marker Mapper\cite{munoz-salinas_mapping_2018} and ArUco\cite{Aruco2014} is that we achieve a better extrinsic calibration than using the standard calibration methods as previously explained. Furthermore, we think the reason of the inferiority of Apriltag 2 \cite{wang_apriltag_2016} is that the detector is not as good as the Aruco detector as we have seen in our experiments. Additionally, there is no object model used in this approach which could be another explanation of why it performs worse than the Aruco + Marker Mapper combination.

\begin{table}[]
    \centering
    \begin{tabular}{c||c|c||c|c||c|c}
          \multirow{4}{1cm}{\centering Radius (m)} & \multicolumn{2}{c||}{\multirow{3}{2cm}{Our Approach}} & \multicolumn{2}{c||}{\multirow{3}{2cm}{ArUco \cite{Aruco2014} + Marker Mapper \cite{munoz-salinas_mapping_2018}}} & \multicolumn{2}{c}{\multirow{3}{2.3cm}{Apriltag 2 \cite{wang_apriltag_2016}}}  \\
          & \multicolumn{2}{c||}{} & \multicolumn{2}{c||}{} & \multicolumn{2}{c}{} \\
          & \multicolumn{2}{c||}{} & \multicolumn{2}{c||}{} & \multicolumn{2}{c}{} \\
         \cline{2-7} & mm & deg & mm & deg & mm & deg\\
       \hline \hline
         \centering 0.7 & 0.63  &  0.83  & 2.43  & 0.97 & 43.42 & 5.05\\
        \hline
         \centering 0.9 & 0.94  & 0.86  & 2.94  & 1.79 & 33.35 &  3.53\\
        \hline
         \centering 1.1 & 0.94  & 0.65  & 6.00  & 2.23 & 48.74 & 9.79\\
         \hline
         \centering 1.3 & 0.88 & 0.85 & 17.34 & 8.21 & 80.06& 11.37 \\
        
    \end{tabular}
    \caption{Translation and Rotation errors (in millimeters and degrees respectively) of tracking the object using different methods at different distances.}
    \label{tab:tracking_errors}
\end{table}

\section{Conclusions}
\label{sec::conclusions}

This paper has proposed a method that automatically estimates the three-dimensional structure of a set of planar markers (object), the poses of a set of cameras, and the relative pose between the object and the cameras. The input for the proposed method is a synchronized video sequence of the object moving freely in front of the cameras. The uses of the proposed method include tracking of devices such as surgical instruments in augmented reality applications, or robot navigation. This is the first work that solves all of these problems simultaneously, up to our knowledge.

The proposed method starts by obtaining initial estimations of the camera poses by creating a connection graph analyzing the marker detected in the video sequence. Then, another graph representing the structure of the markers is created from their observations in the individual cameras. The graphs are employed to obtain initial solutions for both the camera and the object poses that are refined using a global non-linear optimization. The optimized configurations can be employed afterward for tracking purposes.

The proposed method has been evaluated in several experiments and the accuracy measured with an infrared-based motion capture system. The proposed method achieves sub-millimeter position accuracy and sub-degree orientation accuracy in the estimation of object pose even using low-resolution cameras. In addition, the proposed method is fast enough to allow real-time performance using five cameras and a single CPU thread.

\section*{Acknowledgments}
This project has been funded under projects  IFI16/00033 (ISCIII) and TIN2016-75279-P of Spain Ministry of Economy, Industry and Competitiveness and FEDER. The authors thank the Health Time Radiology Company for its help for the Project IFI16/00033 (ISCIII).

\bibliographystyle{ieeetr}
\bibliography{references}

\end{document}